\documentclass[10pt, conference, compsocconf]{IEEEtran}

\usepackage{graphicx}
\usepackage[utf8]{inputenc} 
\usepackage[T1]{fontenc}    
\usepackage{hyperref}       
\usepackage{url}            
\usepackage{booktabs}       
\usepackage{amsfonts}       
\usepackage{nicefrac}       
\usepackage{microtype}      
\usepackage{lipsum}

\begin{document}

\title{Towards Task-Oriented Dialogue in Mixed Domains}

\author{
  \IEEEauthorblockN{Tho Luong Chi}
  \IEEEauthorblockA{
    FPT Technology Research Institute\\
    FPT University, Hanoi, Vietnam\\
    \textit{tholc2@fpt.com.vn}}
  \and
  \IEEEauthorblockN{Phuong Le-Hong}
  \IEEEauthorblockA{
    FPT Technology Research Institute\\
    Vietnam National University, Hanoi, Vietnam\\
    \textit{phuonglh@vnu.edu.vn}}
}

\maketitle

\begin{abstract}
  This work investigates the task-oriented dialogue problem in
  mixed-domain settings. We study the effect of alternating between
  different domains in sequences of dialogue turns using two related
  state-of-the-art dialogue systems. We first show that a specialized state
  tracking component in multiple domains plays an important role and
  gives better results than an end-to-end task-oriented dialogue
  system. We then propose a hybrid system which is able to improve the
  belief tracking accuracy of about 28\% of average absolute point on
  a standard multi-domain dialogue dataset. These experimental results
  give some useful insights for improving our commercial chatbot platform
  FPT.AI, which is currently deployed for many practical chatbot
  applications.
\end{abstract}


\section{Introduction}
\label{sec:introduction}

In this work, we investigate the problem of task-oriented dialogue in
mixed-domain settings. Our work is related to two lines of research in
Spoken Dialogue System (SDS), namely \textit{task-oriented dialogue
  system} and \textit{multi-domain dialogue system}. We briefly review
the recent literature related to these topics as follows.

Task-oriented dialogue systems are computer programs which can
assist users to complete tasks in specific domains by understanding
user requests and generating appropriate responses within several
dialogue turns. Such systems are useful in domain-specific chatbot
applications which help users find a restaurant or book a
hotel. Conventional approach for building a task-oriented dialogue
system is concerned with building a quite complex pipeline of many
connected components. These components are usually independently
developed which include at least four crucial modules: a natural language understanding
module, a dialogue state tracking module, a dialogue policy learning
module, and a answer generation module. Since these systems
components are usually trained independently, their optimization
targets may not fully align with the overall system evaluation
criteria~\cite{Liu:2018}. In addition, such a pipeline system often
suffers from error propagation where error made by upstream modules
are accumuated and got amplified to the downstream ones.

To overcome the above limitations of pipeline task-oriented dialogue
systems, much research has focused recently in designing end-to-end
learning systems with neural network-based models. One key property
of task-oriented dialogue model is that it is required to reason and plan over
multiple dialogue turns by aggregating useful information during the
conversation. Therefore, sequence-to-sequence models such as the
encoder-decoder based neural network models are proven to be
suitable for both task-oriented and non-task-oriented
systems. Serban et al. proposed to build end-to-end dialogue
systems using generative hierarchical recurrent encoder-decoder neural
network~\cite{Serban:2016}. Li et al. presented persona-based models which
incorporate background information and speaking style of interlocutors
into LSTM-based seq2seq network so as to improve the modeling of
human-like behavior~\cite{Li:2016}. Wen et al. designed an end-to-end trainable
neural dialogue model with modularly connected
components~\cite{Wen:2017}. Bordes et al.~\cite{Bordes:2017} proposed a task-oriented dialogue
model using end-to-end memory networks. At the same time, many works
explored different kinds of networks to model the dialogue state, such
as copy-augmented networks~\cite{Eric:2017}, gated memory
networks~\cite{Liu:2017}, query-regression
networks~\cite{Seo:2016}. These systems do not perform slot-filling or
user goal tracking; they rank and select a response from a set of
response candidates which are conditioned on the dialogue history. 

One of the significant effort in developing end-to-end task-oriented
systems is the recent Sequicity framework~\cite{Lei:2018}. This
framework also relies on the sequence-to-sequence model and can be
optimized with supervised or reinforcement learning. The Sequicity
framework introduces the concept of \textit{belief span} (bspan),
which is a text span that tracks the dialogue states at each turn. In
this framework, the task-oriented dialogue problem is decomposed into
two stages: bspan generation and response generation. This framework
has been shown to significantly outperform state-of-the-art
pipeline-based methods. 

\begin{figure*}[t]
  \centering
  \includegraphics[scale=0.4]{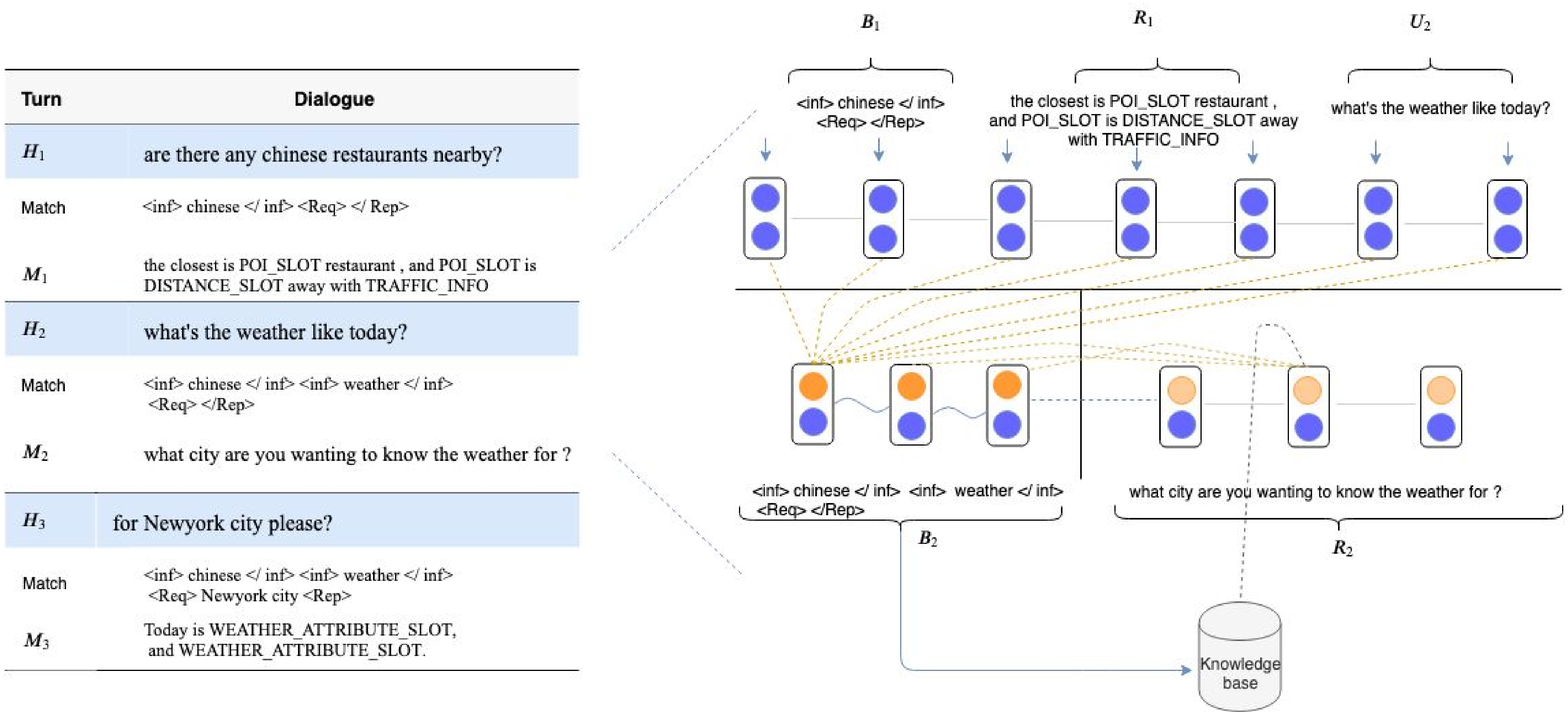}
  \caption{Sequicity architecture.}
  \label{fig:fig1}
\end{figure*}

The second line of work in SDS that is related to this work is
concerned with multi-domain dialogue systems. As presented above, one
of the key components of a dialogue system is dialogue state tracking,
or belief tracking, which maintains the states of conversation. A
state is usually composed of user's goals, evidences and information
which is accumulated along the sequence of dialogue turns. While the
user's goal and evidences are extracted from user's utterances, the
useful information is usually aggregated from external resources such
as knowledge bases or dialogue ontologies. Such knowledge bases
contain slot type and slot value entries in one or several predefined
domains. Most approaches have difficulty scaling up with multiple
domains due to the dependency of their model parameters on the
underlying knowledge bases. Recently, Ramadan et
al.~\cite{Ramadan:2018} has introduced a novel approach which utilizes
semantic similarity between dialogue utterances and knowledge base
terms, allowing the information to be shared across domains. This
method has been shown not only to scale well to multi-domain
dialogues, but also outperform existing state-of-the-art models in
single-domain tracking tasks. 

The problem that we are interested in this work is task-oriented
dialogue in mixed-domain settings. This is different from the
multi-domain dialogue problem above in several 
aspects, as follows:
\begin{itemize}
\item First, we investigate the phenomenon of alternating between
  different dialogue domains in subsequent dialogue turns, where each
  turn is defined as a pair of user question and machine answer. That
  is, the domains are mixed between turns. For example, in the first
  turn, the user requests some information of a restaurant; then in
  the second turn, he switches to the a different domain, for example,
  he asks about the weather at a specific location. In a next turn, he
  would either switch to a new domain or come back to ask about some
  other property of the suggested restaurant. This is a realistic
  scenario which usually happens in practical chatbot applications in
  our observations.  We prefer calling this problem mixed-domain
  dialogue rather than multiple-domain dialogue. 
\item Second, we study the effect of the mixed-domain setting in the
  context of multi-domain dialogue approaches to see how they perform
  in different experimental scenarios.
\end{itemize}

The main findings of this work include:
\begin{itemize}
\item A specialized state tracking component in multiple domains
  still plays an important role and gives better results than a
  state-of-the-art end-to-end task-oriented dialogue system.
\item A combination of specialized state tracking system and an
  end-to-end task-oriented dialogue system is beneficial in mix-domain
  dialogue systems. Our hybrid system is able to improve the belief
  tracking accuracy of about 28\% of average absolute point on a standard
  multi-domain dialogue dataset.
\item These experimental results give some useful insights on data
  preparation and acquisition in the development of the chatbot
  platform FPT.AI\footnote{\url{http://fpt.ai/}}, which is currently
  deployed for many practical chatbot applications.
\end{itemize}

The remainder of this paper is structured as follows. First,
Section~\ref{sec:methodology} discusses briefly the two methods in
building dialogue systems that our method relies on. Next,
Section~\ref{sec:experiments} presents experimental settings and
results. Finally, Section~\ref{sec:conclusion} concludes the paper and
gives some directions for future work.

\section{Methodology}
\label{sec:methodology}

 \begin{figure*}[th]
  \centering
  \includegraphics[scale=0.6]{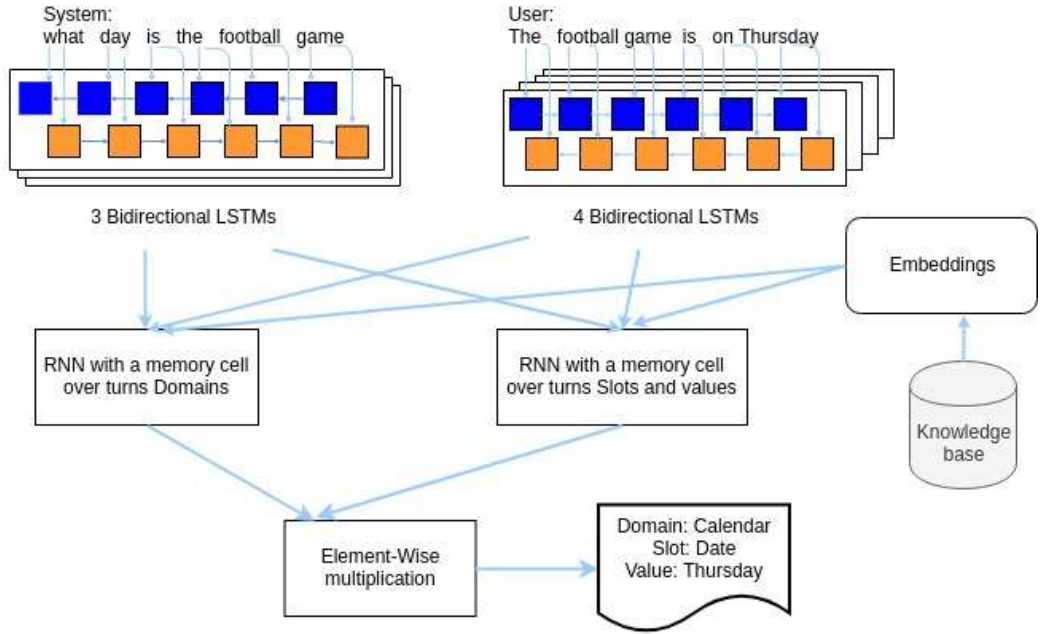}
  \caption{Multi-domain belief tracking with knowledge sharing.}
  \label{fig:fig2}
\end{figure*}

In this section, we present briefly two methods that we use in our
experiments which have been mentioned in the previous section. The
first method is the Sequicity framework and the second one is the
state-of-the-art multi-domain dialogue state tracking approach.

\subsection{Sequicity}

Figure~\ref{fig:fig1} shows the architecture of the Sequicity
framework as described in~\cite{Lei:2018}. In essence,
in each turn, the Sequicity model first takes a bspan ($B_1$) and a
response ($R_1$) which are determined in the previous step, and the current
human question ($U_2$) to generate the current bspan. This bspan is
then used together with a knowledge base to generate the corresponding
machine answer ($R_2$), as shown in the right part of
Figure~\ref{fig:fig1}. 

The left part of that figure shows an example dialogue in a
mixed-domain setting (which will be explained in
Section~\ref{sec:experiments}). 

\subsection{Multi-domain Dialogue State Tracking}

Figure~\ref{fig:fig2} shows the architecture of the multi-domain
belief tracking with knowledge sharing as described
in~\cite{Ramadan:2018}. This is the state-of-the-art belief tracker
for multi-domain dialogue.

This system encodes system responses with 3 bidirectional LSTM network
and encodes user utterances with 3+1 bidirectional LSTM network. There
are in total 7 independent LSTMs. For tracking domain, slot and value,
 it uses 3 corresponding LSTMs, either for system response or user
 utterance. There is one special LSTM to track the user
 affirmation. The semantic similarity between the utterances and
 ontology terms are learned and shared between domains through their
 embeddings in the same semantic space. 

\section{Experiments}
\label{sec:experiments}

In this section, we present experimental settings, different scenarios
and results. We first present the datasets, then implementation
settings, and finally obtained results. 

\subsection{Datasets}

We use the publicly available dataset KVRET~\cite{Eric:2017} in our
experiments. This dataset is created by the Wizard-of-Oz method~\cite{Kelly:1984} on Amazon
Mechanical Turk platform. This dataset includes dialogues in 3 domains: calendar,
weather, navigation (POI) which is suitable for our mix-domain
dialogue experiments. There are 2,425 dialogues for training, 302 for
validation and 302 for testing, as shown in the upper half of Table~\ref{tab:datasets}.

\begin{table*}
  \centering
  \caption{Some statistics of the datasets used in our experiments. The
    original KVRET dataset is shown in the upper half of the table. The mixed
    dataset is shown in the lower half of the table.} \label{tab:datasets}
  \begin{tabular}{lllllc}
    &                       &                                    &                          & \multicolumn{2}{c}{}             \\ \cline{1-4}
    \multicolumn{1}{|l|}{Dataset} & \multicolumn{3}{l|}{\textbf{KVRET}}                                                           & \multicolumn{2}{l}{}                              \\ \cline{1-4}
    \multicolumn{1}{|l|}{Dialogues}    & \multicolumn{3}{l|}{Train: 2,425
                                    ; Test: 302 ;
                                    Dev.: 302}                                        & \multicolumn{2}{l}{}                              \\ \cline{1-4}
    \multicolumn{1}{|l|}{Domains} & \multicolumn{1}{c|}{calendar} &
                                                                    \multicolumn{1}{c|}{weather} & \multicolumn{1}{c|}{POI} & \multicolumn{2}{l}{}                              \\ \cline{1-4}
    &                               &                                    &                          & \multicolumn{2}{l}{}                              \\ \cline{1-4}
    \multicolumn{1}{|l|}{Dataset} & \multicolumn{3}{l|}{\textbf{MIXED DOMAINS}}                                                   & \multicolumn{2}{l}{}                              \\ \hline
    \multicolumn{1}{|l|}{Domains} & \multicolumn{1}{c|}{calendar} & \multicolumn{1}{c|}{weather} & \multicolumn{1}{c|}{POI} & \multicolumn{2}{c|}{Mixed weather -- POI}          \\ \hline
    \multicolumn{1}{|l|}{Train}   & \multicolumn{1}{c|}{828}      & \multicolumn{1}{c|}{398}           & \multicolumn{1}{c|}{400} & \multicolumn{2}{c|}{400}                          \\ \hline
    \multicolumn{1}{|l|}{Test}    & \multicolumn{1}{c|}{102}      & \multicolumn{1}{c|}{50}            & \multicolumn{1}{c|}{50}  & \multicolumn{2}{c|}{50}                           \\ \hline
    \multicolumn{1}{|l|}{Dev.}     & \multicolumn{1}{c|}{102}      & \multicolumn{1}{c|}{50}            & \multicolumn{1}{c|}{50}  & \multicolumn{2}{c|}{50}                  \\ \hline
  \end{tabular}
\end{table*}

In this original dataset, each dialogue is of a single domain where
all of its turns are on that domain. Each turn is composed of a
sentence pair, one sentence is a user utterance, the other sentence is
the corresponding machine response. A dialogue is a sequence of turns. To create
mix-domain dialogues for our experiments, we make some changes in this
dataset as follows:
\begin{itemize}
\item We keep the dialogues in the calendar domain as they are.
\item We take a half of dialogues in the weather domain and a half of
  dialogues in the POI domain and mix their turns together, resulting
  in a dataset of mixed weather-POI dialogues. In this mixed-domain
  dialogue, there is a turn in the weather domain, followed by a turn
  in POI domain or vice versa. 
\end{itemize}
We call this dataset \textit{the sequential turn dataset}. Since the start turn of a dialogue has a special role in triggering
the learning systems, we decide to create another and different
mixed-domain dataset with the following mixing method:
\begin{itemize}
\item The first turn and the last turn of each dialogue are kept as in their original. 
\item The internal turns are mixed randomly.
\end{itemize}
We call this dataset \textit{the random turn dataset}. Some statistics
of these mixed-domain datasets are shown in the lower half of the Table~\ref{tab:datasets}.

\subsection{Experimental Settings}

For the task-oriented Sequicity model, we keep the best parameter settings
as reported in the original framework, on the same KVRET
dataset~\cite{Lei:2018}. In particular, the hidden size of GRU unit is
set to 50; the learning rate of Adam optimizer is 0.003. In addition
to the original GRU unit, we also re-run this framework with simple RNN unit to
compare the performance of different recurrent network types. The
Sequicity tool is freely available for download.\footnote{\url{https://github.com/WING-NUS/sequicity}}

For the multi-domain belief tracker model, we set the hidden size of
LSTM units to 50 as in the original model; word embedding size is 300
and number of training epochs is 100. The corresponding tool is also
freely available for download.\footnote{\url{https://github.com/osmanio2/multi-domain-belief-tracking}}

\subsection{Results}

Our experimental results are shown in Table~\ref{tab:results}. The
first half of the table contains results for task-oriented dialogue
with the Sequicity framework with two scenarios for training data
preparation. For each experiment, we run our models for 3 times and
their scores are averaged as the final score. The \textit{mixed
  training} scenario performs the mixing of both the training data,
development data and the test data as described in the previous
subsection. The \textit{non-mixed training} scenario performs the
mixing only on the development and test data, keeps the training data
unmixed as in the original KVRET dataset. As in the Sequicity
framework, we report entity match rate, BLEU score and Success F1
score. \textbf{Entity match rate} evaluates task completion, it
determines if a system can generate all correct constraints to search
the indicated entities of the user. \textbf{BLEU} score evaluates the
language quality of generated responses. \textbf{Success F1} balances
the recall and precision rates of slot answers. For further details on
these metrics, please refer to~\cite{Lei:2018}.

\begin{table*}
  \centering
  \caption{Our experimental results. \textbf{Match.} and \textbf{Succ. F1} are Entity match rate
    and Success F1. The upper half of the table shows results of
    task-oriented dialogue with the Sequicity framework. The lower half
    of the table shows results of multi-domain belief tracker.}
  \label{tab:results}
  \begin{tabular}{lccccllc}
          \cline{3-8}
          & \multicolumn{1}{l|}{}             &
                      \multicolumn{3}{l|}{Case
         1 - sequential turn}
          & \multicolumn{3}{c|}{Case 2 - random turn }                       \\ \hline
          \multicolumn{2}{|l|}{\textbf{Sequicity}}             & \multicolumn{1}{l|}{Match.}                                                       & \multicolumn{1}{l|}{BLEU}                                                       & \multicolumn{1}{l|}{Succ. F1}                  & \multicolumn{1}{c|}{Match.} & \multicolumn{1}{c|}{BLEU}   & \multicolumn{1}{c|}{Succ. F1} \\ \hline
          \multicolumn{1}{|p{3cm}|}{mixed training}           & \multicolumn{1}{c|}{GRU}          & \multicolumn{1}{c|}{0.6367}                                                       & \multicolumn{1}{c|}{\textbf{0.1930}}                                                     & \multicolumn{1}{c|}{\textbf{0.7358}}                                                     & \multicolumn{1}{l|}{0.6860} & \multicolumn{1}{l|}{0.1862} & \multicolumn{1}{c|}{\textbf{0.7562}}   \\ \cline{2-8} 
          \multicolumn{1}{|l|}{}                                     & \multicolumn{1}{c|}{RNN}          & \multicolumn{1}{c|}{0.7354}                                                       & \multicolumn{1}{c|}{0.1847}                                                     & \multicolumn{1}{c|}{0.7129}                                                     & \multicolumn{1}{l|}{0.6591} & \multicolumn{1}{c|}{0.1729} & \multicolumn{1}{c|}{0.7105}   \\ \hline
          \multicolumn{1}{|p{3cm}|}{non-mixed training} & \multicolumn{1}{c|}{GRU}          & \multicolumn{1}{c|}{0.7399}                                                       & \multicolumn{1}{c|}{0.1709}                                                     & \multicolumn{1}{c|}{0.7055}                                                     & \multicolumn{1}{c|}{\textbf{0.7488}} & \multicolumn{1}{c|}{0.1820} & \multicolumn{1}{c|}{0.7173}   \\ \cline{2-8} 
          \multicolumn{1}{|l|}{}                                     & \multicolumn{1}{c|}{RNN}          & \multicolumn{1}{c|}{0.7706}                                                       & \multicolumn{1}{c|}{0.1453}                                                     & \multicolumn{1}{c|}{0.6156}                                                     & \multicolumn{1}{c|}{0.6995} & \multicolumn{1}{c|}{0.1580} & \multicolumn{1}{c|}{0.6633}   \\ \hline
          & \multicolumn{1}{l|}{}             & \multicolumn{1}{l|}{\begin{tabular}[c]{@{}l@{}}Domain - \\ accuracy\end{tabular}} & \multicolumn{1}{c|}{\begin{tabular}[c]{@{}c@{}}Slot - \\ accuracy\end{tabular}} & \multicolumn{1}{c|}{\begin{tabular}[c]{@{}c@{}}Value -\\ accuracy\end{tabular}} & \multicolumn{1}{l|}{}       & \multicolumn{1}{l|}{}       & \multicolumn{1}{l|}{}         \\ \hline
          \multicolumn{1}{|l|}{\textbf{Belief tracker}}      & \multicolumn{1}{c|}{Multi-domain} & \multicolumn{1}{c|}{0.8253}                                                       & \multicolumn{1}{c|}{\textbf{0.9329}}                                                     & \multicolumn{1}{c|}{\textbf{0.9081}}                                                     & \multicolumn{1}{l|}{}       & \multicolumn{1}{l|}{}       & \multicolumn{1}{l|}{}         \\ \cline{2-8} 
          \multicolumn{1}{|l|}{}                                     & \multicolumn{1}{c|}{Sequicity}    & \multicolumn{1}{l|}{}                                                             & \multicolumn{1}{c|}{0.7171}                                                     & \multicolumn{1}{c|}{0.5644}                                                     & \multicolumn{1}{l|}{}       & \multicolumn{1}{l|}{}       & \multicolumn{1}{l|}{}         \\ \hline
          & \multicolumn{1}{l}{}              & \multicolumn{1}{l}{}                                                              & \multicolumn{1}{l}{}                                                            & \multicolumn{1}{l}{}                                                            &                             &                             & \multicolumn{1}{l}{}         
        \end{tabular}
 \end{table*}
      
 In the first series of experiments, we evaluate the Sequicity
 framework on different mixing scenarios and different recurrent units
 (GRU or RNN), on two mixing methods (sequential turn or random turn),
 as described previously. We see that when the training data is kept
 unmixed, the match rates are better than those of the mixed training
 data. It is interesting to note that the GRU unit is much more
 sensitive with mixed data than the simple RNN unit with the
 corresponding absolute point drop of about 10\%, compared to about
 3.5\%. However, the entity match rate is less important than the
 Success F1 score, where the GRU unit outperforms RNN in both
 sequential turn and random turn by a large margin. It is logical that
 if the test data are mixed but the training data are unmixed, we get
 lower scores than when both the training data and test data are
 mixed. The GRU unit is also better than the RNN unit on response
 generation in terms of BLEU scores.

 We also see that the task-oriented dialogue system has difficulty
 running on mixed-domain dataset; it achieves only about 75.62\% of
 Success F1 in comparison to about 81.1\% (as reported in the
 Sequicity paper, not shown in our table). Appendix~\ref{sec:examples}
 shows some example dialogues generated automatically by our
 implemented system.

 In the second series of experiments, we evaluate the belief tracking
 components of two systems, the specialized multi-domain belief
 tracker and the Sequicity bspan component. As shown in the lower half
 of the Table~\ref{tab:results}, Sequicity capability of belief
 tracking is much worse than that of the multi-domain belief
 tracker. The slot accuracy gap between the tools is about 21.6\%, the
 value accuracy gap is about 34.4\%; that is a large average gap of
 28\% of accuracy. This result suggests a future work on combining a 
 specialized belief tracking module with an end-to-end task-oriented
 dialogue system to improve further the performance of the overall
 dialogue system.

\subsection{Error Analysis}
In this subsection, we present an example of erroneous mixed dialogue with multple turns. Table~\ref{tab:error}
shows a dialogue in the test set where wrong generated responses of the Sequicity system are marked in bold font.

\begin{table*}
  \centering
  \caption{A mixed dialogue example in the test set with erroneous generated responses. The last two columns show
    respectively the system's generated bspan and the gold bspan or belief tracker.}\label{tab:error}
  \begin{tabular}{| l | p{7cm} | l | l | l |}
  \hline
\textbf{Turn}	&	\textbf{Dialogue}	&	\textbf{Domain}	&	\textbf{Generated bspan}	&	\textbf{Gold bspan}\\	\hline
1.Human	&	car please send me the fastest route to go back home	&	POI	&	EOS\_Z1	&	home fastest EOS\_Z1\\	\hline
1.Machine	&	there is \textbf{heavy traffic} on the way to \textbf{Pizza Hut}.	&	POI	&		&	\\	\hline
2.Human	&	what is the weather like this week	&	Weather info	&	EOS\_Z1	&	this\_week EOS\_Z1 \\ \hline
2.Machine	&	what city are you interested in ?	&	Weather info	&		&	\\	\hline
3.Human	&	Cleveland .	&	Weather info	&	cleveland EOS\_Z1	&	cleveland EOS\_Z1 \\	\hline
3.Machine	&	\textbf{it will be clear sky on monday} 	&	Weather info	&		&	\\	\hline
4.Human	&	what is the address ?	&	POI	&	cleveland EOS\_Z1	&	home fastest  EOS\_Z1\\	\hline
4.Machine	&	\textbf{cleveland}  is located at \textbf{5672 barringer street}	&	POI	&		&	\\	\hline
5.Human	&	please send me the fastest route on screen and let 's go there , thank you !	&	POI	&	cleveland EOS\_Z1	&	home fastest  EOS\_Z1\\	\hline
5.Machine	&	there is \textbf{moderate traffic} on our way but i sent the route to your screen , drive carefully !	&	POI	&		&	\\	\hline
6.Human	&	thank .	&		&	cleveland EOS\_Z1	&	EOS\_Z1\\	\hline
6.Machine	&	you are welcome .	&		&		&	\\	\hline
\end{tabular}
\end{table*}  

In the first turn, the system predicts incorrectly the bspan, thus generates wrong slot values (\texttt{heavy traffic} and \texttt{Pizza Hut}).
The word \texttt{Pizza Hut} is an arbitrary value selected by the system when it cannot capture the correct value \texttt{home} in the bspan.
In the second turn, the machine is not able to capture the value \texttt{this\_week}. This failure does not manifest immediately at this turn
but it is accumulated to make a wrong answer at the third turn (\texttt{monday} instead of \texttt{this\_week}).

The third turn is of domain weather and the fourth turn is switched to domain POI. The bspan value \texttt{cleveland} is retained through cross domain,
resulting in an error in the fourth turn, where \texttt{cleveland} is shown instead of \texttt{home}. This example demonstrates a weakness of the system
when being trained on a mixed-domain dataset. In the fifth turn, since the system does not recognize the value \texttt{fastest} in the bspan, it generates
a random and wrong value \texttt{moderate traffic}. Note that the generated answer of the sixth turn is correct despite of the wrong predicted bspan; however,
it is likely that if the dialogue continues, this wrong bspan may result in more answer mistakes. In such situations, multi-domain belief tracker usually
performs better at bspan prediction.

\section{Conclusion}
\label{sec:conclusion}

We have presented the problem of mixed-domain task-oriented dialogue
and its empirical results on two datasets. We employ two
state-of-the-art, publicly available tools, one is the Sequicity
framework for task-oriented dialogue, and another is the multi-domain
belief tracking system. The belief tracking capability of the
specialized system is much better than that of the end-to-end
system. We also show the difficulty of task-oriented dialogue systems
on mixed-domain datasets through two series of experiments. These
results give some useful insights in combining the approaches to
improve the performance of a commercial chatbot platform which is
under active development in our company. We plan to extend this
current research and integrate its fruitful results into a future
version of the platform.

\bibliographystyle{unsrt}

\bibliography{references}

\begin{thebibliography}{10}

\bibitem{Liu:2018}
Bing Liu, Gokhan Tur, Dilek Hakkani-Tur, Pararth Shah, and Larry Heck.
\newblock Dialogue learning with human teaching and feedback in end-to-end
  trainable task-oriented dialogue systems.
\newblock In {\em Proceedings of NAACL}, 2018.

\bibitem{Serban:2016}
Iulian Serban, Alessandro Sordoni, Yoshua Bengio, Aaron~C. Courville, and
  Joelle Pineau.
\newblock Building end-to-end dialogue systems using generative hierarchical
  neural network models.
\newblock In {\em Proceedings of AAAI}, 2016.

\bibitem{Li:2016}
Jiwei Li, Michel Galley, Chris Brockett, Georgios~P. Spithourakis, Jianfeng
  Gao, and Bill Dolan.
\newblock A persona-based neural conversation model.
\newblock In {\em Proceedings of ACL}, 2016.

\bibitem{Wen:2017}
Tsung-Hsien Wen, David Vandyke, Nikola Mrksic, Milica Gasic, Lina~M.
  Rojas-Barahona, Pei-Hao Su, Stefan Ultes, and Steve Young.
\newblock A network-based end-to-end trainable task-oriented dialogue system.
\newblock In {\em Proceedings of EACL}, 2017.

\bibitem{Bordes:2017}
Antoine Bordes, Y-Lan Boureau, and Jason Weston.
\newblock Learning end-to-end goal-oriented dialogue.
\newblock In {\em Proceedings of ICLR}, 2017.

\bibitem{Eric:2017}
Mihail Eric and Christopher~D. Manning.
\newblock A copy-augmented sequence-to-sequence architecture gives good
  performance on task-oriented dialogue.
\newblock In {\em Proceedings of EACL}, 2017.

\bibitem{Liu:2017}
Fei Liu and Julien Perez.
\newblock Gated end-to-end memory networks.
\newblock In {\em Proceedings of EACL}, 2017.

\bibitem{Seo:2016}
Min~Joon Seo, Hannaneh Hajishirzi, and Ali Farhadi.
\newblock Query-regression networks for machine comprehension.
\newblock In {\em Preprint ArXiv}, 2016.

\bibitem{Lei:2018}
Wenqiang Lei, Xisen Jin, Zhaochun Ren, Xiangnan He, Min-Yen Kan, and Dawei Yin.
\newblock Sequicity: Simplifying task-oriented dialogue systems with single
  sequence-to-sequence architectures.
\newblock In {\em Proceedings of ACL}, 2018.

\bibitem{Ramadan:2018}
Milica~Gasic Osman~Ramadan, Paweł~Budzianowski.
\newblock Large-scale multi-domain belief tracking with knowledge sharing.
\newblock In {\em Proceedings of ACL}, 2018.

\bibitem{Kelly:1984}
John~F Kelley.
\newblock iterative design methodology for user-friendly natural language
  office information applications.
\newblock {\em ACM Transactions on Information Systems}, 2(1):26--41, 1984.

\end{thebibliography}

\newpage
\appendix
\section{Example Dialogues}
\label{sec:examples}

The following is three example dialogues generated by our system. The
first dialogue is in single-domain.

\begin{center}
  \begin{tabular}{lp{4cm}l}
    \hline
    \textbf{Turn} & \textbf{Dialogue 1} & \textbf{Domain} \\ \hline
    Human         & What time is my next dinner event and who will be attending? & Calendar    \\ \hline
    {Machine}       & {your dinner is on MONDAY at 5 PM .} & {Calendar} \\ \hline
    {Human}         & {That will do, thanks.} &  \\ \hline
    {Machine}       & {you are welcome} & \\ \hline
  \end{tabular}
\end{center}

The next two dialogues are in mixed-domains.

\begin{center}
  \begin{tabular}{lp{4cm}l}
    \hline
    \textbf{Turn} & \textbf{Dialogue 2} & {\textbf{Domain}} \\ \hline
    {Human}         & Please give me the address and directions to home within 5 miles of here, avoiding all heavy traffic. & POI\\ \hline
    {Machine}       & {HOME is located at 5671 BARRINGER STREET}                                                            & POI             \\ \hline
    {Human}         & {what is the forecast for the next 7 days?}                                             & {weather info}  \\ \hline
    {Machine}  & {what city are you interested in ?} & {weather info}\\ \hline
    {Human} & {Cleveland, will it be windy?} & {weather info}    \\ \hline
    {Machine} & {it will not be windy in Cleveland this week .} & {weather info}\\ \hline
    {Human} & {Thank you.} & \\ \hline
    {Machine} & {you’re welcome.} &\\ \hline
  \end{tabular}
\end{center}

\begin{center}
  \begin{tabular}{lp{4cm}l}
    \hline
    {\textbf{Turn}} & {\textbf{Dialogue 3}} &{\textbf{Domain}} \\ \hline
    {Human}& are there any chinese restaurants nearby? & POI\\ \hline
    {Machine} & the closest is TAI PAN restaurant , and TAI PAN
                is 5 miles away with NO TRAFFIC& POI\\ \hline
    {Human}& what's the weather like today? &{weather info} \\ \hline
    Machine& what city are you wanting to know the weather for ? & {weather info}\\ \hline
    Human & for Newyork city please?{weather info} \\ \hline
    Machine & Today is CLEAR SKIES, and WARM. &{weather info}    \\ \hline
    Human & ok, good, set the direction to there & {POI} \\ \hline
    Machine & the address is 830 Almanor Ln , i sent it on your screen &{POI} \\ \hline
    Human & ok thanks&\\ \hline
    Machine & you're welcome ! & \\ \hline
  \end{tabular}
 \end{center}

\end{document}